\documentclass{edm_template}

\begin{document}

\title{GritNet: Student Performance Prediction \\ with Deep Learning}
\numberofauthors{1} 
\author{
\alignauthor Byung-Hak Kim, Ethan Vizitei, Varun Ganapathi\\
       \affaddr{Udacity}\\
       \affaddr{2465 Latham Street}\\
       \affaddr{Mountain View, CA 94040}\\
       \email{\{hak, ethan, varun\}@udacity.com}
}
\maketitle
\begin{abstract}
Student performance prediction - where a machine forecasts the future performance of students as they interact with online coursework - is a challenging problem. Reliable early-stage predictions of a student's future performance could be critical to facilitate timely educational interventions during a course. However, very few prior studies have explored this problem from a deep learning perspective. In this paper, we recast the student performance prediction problem as a sequential event prediction problem and propose a new deep learning based algorithm, termed GritNet, which builds upon the bidirectional long short term memory (BLSTM). Our results, from real Udacity students' graduation predictions, show that the GritNet not only consistently outperforms the standard logistic-regression based method, but that improvements are substantially pronounced in the first few weeks when accurate predictions are most challenging.
\end{abstract}

\keywords{Student performance prediction, Deep learning for education, Educational data mining, Learning analytics}

\section{Introduction}
\label{introduction}
Education is no longer a one-time event but a lifelong experience. One reason is that working lives are now so lengthy and fast-changing that people need to keep learning throughout their careers \cite{economist17}. While the classic model of education is not scaling to meet these changing needs, the wider market is innovating to enable workers to learn in new ways. Massive open online courses (MOOCs), offered by companies such as Udacity and Coursera, are now focusing much more directly on courses that make their students more employable. At Coursera and Udacity, students pay for short programs that bestow \textit{microcredentials} and \textit{Nanodegrees} in technology-focused subjects such as self-driving cars and Android. Moreover, universities are offering online degrees to make it easier for professionals to access opportunities to develop their skills (e.g., Georgia Tech's Computer Science Master's degree). 

However, broadening access to cutting-edge vocational subjects does not naturally guarantee student success \cite{isaac16}\footnote{HarvardX and MITx have reported that only 5.5\% of people who enroll in one of their online courses earn a certificate.}. In a classic classroom, where student numbers are limited, various dimensions of interactions enable the teacher to quite effectively assess an individual student's level of engagement, and anticipate their learning outcomes (e.g., successful completion of a course, course withdrawals, final grades). In the world of MOOCs, the significant increase in student numbers makes it impractical for even experienced \textit{human} instructors to conduct such individual assessments. An automated system, which accurately predicts how students will perform in real-time, could possibly help in this case. It would be a valuable tool for making smart decisions about when to make live educational interventions during the course (and with whom), with the aim of increasing engagement, providing motivation and empowering students to succeed.

The student performance prediction problem has been partly studied within the learning analytics and educational data mining communities in the form of the student dropout (or completion) prediction problem (which is an important subclass problem of the student performance prediction problem). Most previous works can be divided into two approaches:
\begin{itemize}
\item The first traditional approach principally relies on generalized linear models, including logistic regression, linear SVMs and survival analysis (see \cite{whitehill17} for a thorough summary). Each model considers different types of behavioral and predictive features extracted from various raw activity records (e.g., clickstream, grades, forum, grades). 
\item The second emerging approach involves an exploration of neural networks (NN). Few prior works explore deep neural network (DNN) model \cite{whitehill17}, recurrent neural network (RNN) model \cite{mi15} and convolutional neural networks (CNN) followed by RNN \cite{wang17}. However, all of these new models, so far, have shown primitive performance. This is mainly because the models still rely on feature engineering to reduce input dimensions which appears to limit one to develop larger (i.e., better) NN models.
\end{itemize}

Student activity records collected from different courses often have various lengths, formats and content, so that features that are effective in one course might not be so in another. Even carefully designed feature dimensions are usually constrained to be small\footnote{In past works, DNN of width 5 \cite{whitehill17} and LSTM of 20 cell dimensions \cite{mi15} are used.}. Both of these deficiencies produce inputs that are, so far, too restricted to tap the full benefits of sequential deep learning models. To avoid the deficiencies of prior works, GritNet takes students' learning activities across time as raw input (see Section~\ref{Input Representation}) and (implicitly) searches for parts of an event embedding sequence that are most discrminative to predicting a student's performance without having to engineer those parts as an (explicit) input feature (see Section~\ref{Model Architecture}).

In the remainder of this paper, we introduce the basic GritNet model in Section~\ref{GritNet}, followed by the Udacity data and training discussions in Section~\ref{Data and Training}. In Section~\ref{Prediction Performance}, we demonstrate the performance of GritNet via experimental results and give conclusions in Section~\ref{Conclusion}. 

\section{G\lowercase{rit}N\lowercase{et}}
\label{GritNet}
\subsection{Problem Formulation}
\label{ProblemFormulation}
 The task of predicting student performance can be expressed as a sequential event prediction problem \cite{Cynthia11}: given a past event sequence $\mathbf{o}\triangleq(o_{1},\dots,o_{T})$ taken by a student, estimate likelihood of future event sequence $\mathbf{y}\triangleq(y_{T+D},\dots,y_{T'})$ where $D \in \mathbb Z_{+}$. 
 
 In the form of online classes, each event $o_{t}$ represents a student's action (or activities) associated with a time stamp. In other words, $o_{t}$ is defined as a paired tuple of $(a_{t},d_{t})$. Each action $a_{t}$ represents, for example, ``a lecture video viewed'', ``a quiz answered correctly/incorrectly'', or ``a project submitted and passed/failed'', and $d_{t}$ states the corresponding (logged) time stamp. 
 
 Then, log-likelihood of $p(\mathbf{y}|\mathbf{o})$ can be written as Equation~\ref{eq1}, given fixed-dimensional embedding representation $\upsilon$ of $\mathbf{o}$.
\begin{equation} \label{eq1}
\begin{split}
\log p(\mathbf{y}|\mathbf{o}) & \simeq \sum_{i=T+D}^{T'} \log p(y_{i}|\upsilon)
\end{split}
\end{equation}
The goal of each GritNet is, therefore, to compute an individual log-likelihood $\log p(y_{i}|\upsilon)$, and those estimated scores can be simply added up to estimate long-term student outcomes.

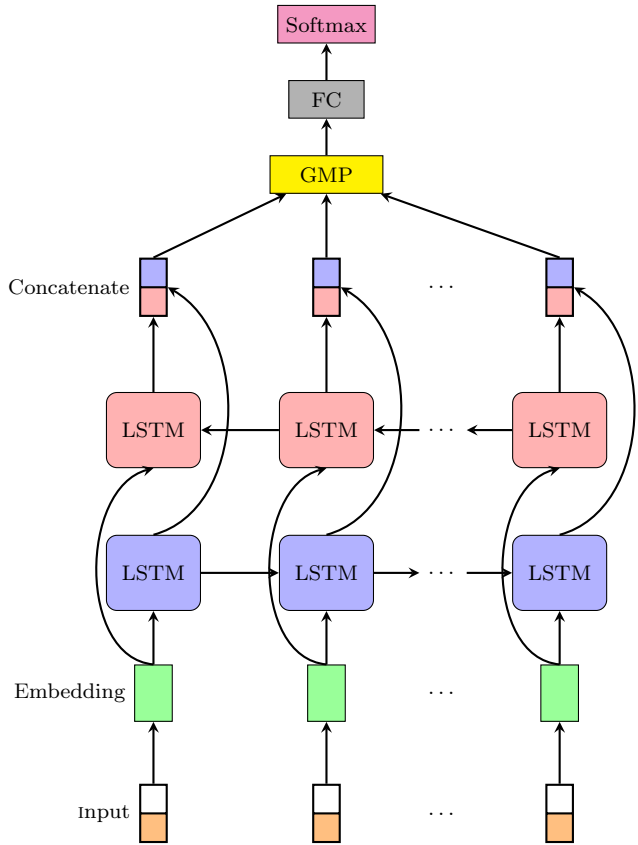
\begin{figure}[ht]
\vskip 0.2in
\usetikzlibrary{shapes.geometric, shapes.multipart, arrows, calc, positioning}

\tikzstyle{bw} = [rectangle, rounded corners, minimum width=1.25cm, minimum height=1cm,text centered, draw=black, fill=red!30]
\tikzstyle{fw} = [rectangle, rounded corners, minimum width=1.25cm, minimum height=1cm,text centered, draw=black, fill=blue!30]
\tikzstyle{emb} = [rectangle, minimum width=0.50cm, minimum height=0.75cm, text centered, draw=black, fill=green!40]
\tikzstyle{feat}=[rectangle split, rectangle split parts=2, rectangle split part fill={white!50, orange!50}, draw=black, thick]
\tikzstyle{cat}=[rectangle split, rectangle split parts=2, rectangle split part fill={blue!30, red!30}, draw=black, thick] 
\tikzstyle{agg} = [rectangle, minimum width=1.5cm, minimum height=0.5cm,text centered, draw=black, fill=yellow!100]
\tikzstyle{fc} = [rectangle, minimum width=1.0cm, minimum height=0.5cm,text centered, draw=black, fill=black!30]
\tikzstyle{softmax} = [rectangle, minimum width=1.0cm, minimum height=0.5cm,text centered, draw=black, fill=magenta!50]

\tikzstyle{arrow} = [thick,->,>=stealth]

\begin{tikzpicture}[node distance=1.5cm]

    \tikzstyle{every node}=[font=\small]

    \node (cat0) [cat, label={left:\textsc{C}oncatenate}]{};
    \node (cat1) [cat, right of=cat0, xshift=0.8cm] {};
    \node (cat2) [cat, right of=cat1, xshift=1.6cm] {};
    \node (dot_cat) at ($(cat1)!.5!(cat2)$) {\ldots};
    
    \node (agg) [agg, above of=cat1] {GMP};
    \node (fc) [fc, above of=agg, yshift=-0.5cm] {FC};
    \node (softmax) [softmax, above of=fc, yshift=-0.5cm] {Softmax};
    
    \node (bw0) [bw, below of=cat0, yshift=-0.4cm] {LSTM};
    \node (bw1) [bw, right of=bw0, xshift=0.8cm] {LSTM};
    \node (bw2) [bw, right of=bw1, xshift=1.6cm] {LSTM};
    \node (dot_bw) at ($(bw1)!.5!(bw2)$) {\ldots};
    
    \node (fw0) [fw, below of=bw0, yshift=-0.4cm] {LSTM};
    \node (fw1) [fw, right of=fw0, xshift=0.8cm] {LSTM};
    \node (fw2) [fw, right of=fw1, xshift=1.6cm] {LSTM};
    \node (dot_fw) at ($(fw1)!.5!(fw2)$) {\ldots};
    
    \node (emb0) [emb, below of=fw0, yshift=-0.1cm, label={left:\textsc{E}mbedding}] {};
    \node (emb1) [emb, right of=emb0, xshift=0.8cm] {};
    \node (emb2) [emb, right of=emb1, xshift=1.6cm] {};
    \node (dot_emb) at ($(emb1)!.5!(emb2)$) {\ldots};

    \node (feat0) [feat, below of=emb0, yshift=-0.1cm, label={left:\textsc{i}nput}] {};
    \node (feat1) [feat, right of=feat0, xshift=0.8cm] {};
    \node (feat2) [feat, right of=feat1, xshift=1.6cm] {};
    \node (dot_feat) at ($(feat1)!.5!(feat2)$) {\ldots};
    
    \draw [arrow] (agg) -- (fc);
    \draw [arrow] (fc) -- (softmax);
    
    \draw [arrow] (cat0.north) -- (agg);
    \draw [arrow] (cat1.north) -- (agg);
    \draw [arrow] (cat2.north) -- (agg);
    
    \draw [arrow] (bw0) -- (cat0);
    \draw [arrow] (bw1) -- (cat1);
    \draw [arrow] (bw2) -- (cat2);
    
    \draw [arrow] (bw2) -- (dot_bw);
    \draw [arrow] (dot_bw) -- (bw1);
    \draw [arrow] (bw1) -- (bw0);
    
    \draw [arrow] (fw0) -- (fw1);
    \draw [arrow] (fw0.north) to [out=15,in=-30] (cat0.east);
    \draw [arrow] (fw1) -- (dot_fw);
    \draw [arrow] (fw1.north) to [out=15,in=-30] (cat1.east);
    \draw [arrow] (dot_fw) -- (fw2);
    \draw [arrow] (fw2.north) to [out=15,in=-30] (cat2.east);
    
    \draw [arrow] (emb0) -- (fw0);
    \draw [arrow] (emb0.north) to [out=175,in=195] (bw0.south);
    \draw [arrow] (emb1) -- (fw1);
    \draw [arrow] (emb1.north) to [out=175,in=195] (bw1.south);
    \draw [arrow] (emb2) -- (fw2);
    \draw [arrow] (emb2.north) to [out=175,in=195] (bw2.south);
    
    \draw [arrow] (feat0) -- (emb0);
    \draw [arrow] (feat1) -- (emb1);
    \draw [arrow] (feat2) -- (emb2);

\end{tikzpicture}
\caption{Architecture of a GritNet for the student performance prediction problem as described in Section \ref{GritNet Architecture}.}
\label{GritNetArch}
\vskip -0.2in
\end{figure}

\subsection{Baseline Model}
\label{BaselineModel}
In order to assess how much added value is brought by the GritNet, logistic regression is used as a baseline model. Here, we use the bag of words (BoW) model to represent each student\mbox{'}s past event sequence $\mathbf{o}$. After transforming all students' activities into a BoW, we count the number of times each unique activity appears in $\mathbf{o}$. 

Let fixed-dimensional feature representation $\mathbf{\upsilon}$ of $\mathbf{o}$ be an $N$-dimensional feature vector where $\upsilon_{j} \in \mathbb Z_{\geq 0}$. Given $\mathbf{\upsilon}$, logistic regression models $\log p(y_{i}|\upsilon)$ as follows:
\begin{equation} \label{eq2}
\begin{split}
\log p(y_{i}=1|\mathbf{\upsilon};\mathbf{\theta})=\frac{1}{1+\exp{( - \mathbf{\theta}^T\mathbf{\upsilon}})},
\end{split}
\end{equation}
where $\mathbf{\theta} \in \mathbb{R}^N$ are the parameters of the logistic regression model. For $M$ training instances $\left \{\big(\mathbf{\upsilon}^{(k)},y^{(k)}\big) \right \}_{k=1}^{M}$, $L_{2}$ regularized logistic regression finds the parameters $\mathbf{\theta}$ that solve the following optimization problem:
\begin{equation} \label{eq3}
\begin{split}
\arg\max_{\mathbf{\theta}} \sum_{k=1}^{m}\log p(y^{(k)}|\mathbf{\upsilon}^{(k)};\mathbf{\theta})+\alpha \|\theta\|_{2}.
\end{split}
\end{equation}
Often, it will be convenient to consider $L_{1}$ regularized logistic regression instead of Equation~\ref{eq2} to handle irrelevant features \cite{Ng04}. We noticed even simpler feature selection methods (e.g., Chi-Square score based), combined with  $L_{2}$ regularized logistic regression, provides  similar results as the $L_{1}$ based.

\subsection{GritNet Architecture}
\label{GritNet Architecture}
\subsubsection{Input Representation}
\label{Input Representation}
In order to feed students' raw event records into the GritNet, it is necessary to encode the time-stamped logs (ordered sequentially) into a sequence of fixed-length input vectors\footnote{GritNet does not need manual feature selections \cite{mi15} or time-series input aggregations per normalized time intervals \cite{wang17}.}. We do this simply by \textit{one-hot encoding}. A one-hot vector $\mathbbm{1}(a_{t}) \in \{0,1\}^L$, where $L$ is the number of unique actions and $j$-th element defined as: 
\begin{equation} \label{eq4}
\begin{split}
\mathbbm{1}(a_{t})_{j}\triangleq
\begin{cases}
    1              & \text{if } j= a_{t}\\
    0              & \text{otherwise}
\end{cases},
\end{split}
\end{equation}
is used to distinguish each activity $a_{t}$ from every other. Then, we connect one-hot vectors of the same student into a long vector sequence to represent the student's whole sequential activities in $\mathbf{o}$. 

While this encoding method preserves ordering information, in contrast to the BoW method (see Section \ref{BaselineModel}), it has a limitation in capturing students' learning speed. Varying learning speed is an important piece of time-dependent information, reflecting a student's progress and/or the course content's difficulty. Since directly employing each time stamp $d_{t}$ will increase the input space too fast, we define the discretized time difference between adjacent events as\footnote{For the Udacity data described in Section~\ref{Udacity Data}, we use day to represent inter-event time intervals.}: 
\begin{equation} \label{eq5}
\begin{split}
\Delta_{t} \triangleq d_{t}-d_{t-1}.
\end{split}
\end{equation}
Then, one-hot encode $\Delta_{t}$ into $\mathbbm{1}(\Delta_{t})$ and connect them with the corresponding $\mathbbm{1}(a_{t})$ to represent $\mathbbm{1}(o_{t})$ as:
\begin{equation} \label{eq6}
\begin{split}
\mathbbm{1}(o_{t}) \triangleq [\mathbbm{1}(a_{t});\mathbbm{1}(\Delta_{t})].
\end{split}
\end{equation}
Lastly, we pre-pad the output sequences shorter than the maximum event sequence length (of a given training set) with all $\mathbf{0}$ vectors.

\begin{figure}[ht]
\twocolumn[
\vskip 0.1in
\begin{tikzpicture}

\begin{axis}[
    xlabel={Weeks into ND Programs}, 
    ylabel={Number of Students},
    xmin=0.5, xmax=8.5, 
    ymin=0, ymax=9000,
    xtick={1,2,...,7,8}, 
    ytick={0,1000,...,8000,9000},
    xmajorgrids=true, 
    ymajorgrids=true, 
    grid style=dashed,
    label style={font=\tiny}, 
    tick label style={font=\tiny},  
    legend pos=south east,
    legend style={font=\tiny}
]

\addplot[color=red,mark=triangle,line width=0.8pt]
    coordinates {(1,1190)
                (2,1292)
                (3,1393)
                (4,1443)
                (5,1543)
                (6,1711)
                (7,1778)
                (8,1819)};

\addplot[color=blue,mark=square,line width=0.8pt]
    coordinates {(1,7542)
                (2,7659)
                (3,7705)
                (4,7932)
                (5,8048)
                (6,8133)
                (7,8188)
                (8,8236)};

\legend{ND-A Dataset, ND-B Dataset}
\end{axis}

\end{tikzpicture}
\begin{tikzpicture}

\begin{axis}[
    xlabel={Weeks into ND Programs}, 
    ylabel={Event Sequence Length},
    xmin=0.5, xmax=8.5, 
    ymin=0, ymax=3500,
    xtick={1,2,...,7,8}, 
    ytick={0,500,...,3000,3500},
    xmajorgrids=true, 
    ymajorgrids=true, 
    grid style=dashed,
    label style={font=\tiny}, 
    tick label style={font=\tiny},  
    legend pos=south east,
    legend style={font=\tiny}
]

\addplot[color=red,mark=triangle,line width=0.8pt]
    coordinates {(1,887)
                (2,1046)
                (3,1171)
                (4,1231)
                (5,1400)
                (6,1449)
                (7,1586)
                (8,1817)};

\addplot[color=blue,mark=square,line width=0.8pt]
    coordinates {(1,1190)
                (2,1223)
                (3,1292)
                (4,1528)
                (5,2066)
                (6,2302)
                (7,2824)
                (8,3255)};

\legend{ND-A Dataset, ND-B Dataset}
\end{axis}

\end{tikzpicture}
\caption{Student and event characteristics of two Udacity Nanodegree program datasets (ND-A and ND-B) by  week: size of dataset (left) and maximum event sequence length (right). The reason the number of students climbs over time is that we only include students who have interacted with their mentor, so if they do not interact in the first couple weeks they are not included early on (left). As a student progresses, the accumulated event sequence gets longer (right).}
\label{Data-figure}
\vskip 0.1in
]
\end{figure}

\subsubsection{Model Architecture}
\label{Model Architecture}
The core of our GritNet model is the embedding \cite{Bengio01}, BLSTM \cite{Alex05} and GMP \cite{James91} layers trained to ingest past student events and predict a log likelihood of a future one. The first embedding layer\footnote{With an embedding layer which provides a dense representation for an event, GritNet achieves an improved performance on the Udacity dataset. Furthermore, after training, similar events appear to be closer in the embedding event space.} learns an embedding matrix $\mathbf{E}^o \in \mathbb{R}^{E\times|O|}$, where $E$ and $|O|$ are the embedding dimension and the number of unique events (i.e., input vector $\mathbbm{1}(o_{t})$ size), to convert an input vector $\mathbbm{1}(o_{t})$ onto a low-dimensional embedding $\mathbf{\upsilon}_{t}$ defined as: 
\begin{equation} \label{eq7}
\begin{split}
\mathbf{\upsilon}_{t} \triangleq \mathbf{E}^o\mathbbm{1}(o_{t}).
\end{split}
\end{equation}

This event embedding $\mathbf{\upsilon}_{t}$ is then passed into the BLSTM and the output vectors are formed by concatenating each forward and backward direction outputs. Next, a GMP layer is added before the output layer. With the GMP layers, GritNet learns to focus the most relevant part of the event embedding sequence while ignoring the rest. This GMP operation seems crucial in boosting prediction power, particularly for imbalanced data provided without any feature engineering\footnote{We empirically find that vanilla BLSTM (without the GMP layer) on the (imbalanced) Udacity datasets does not yield comparable results as shown in Figure~\ref{Performance-figure}. A GMP layer appears to combat this imbalanced data issue effectively by ensuring training errors back-propagate only to the network weights corresponding to the most discriminative part within the event embedding sequence.}. 

The GMP layer output is, ultimately, fed into a fully-connected layer and a softmax (i.e., sigmoid) layer sequentially to calculate the log-likelihood $\log p(y_{i}|\upsilon)$. The complete GritNet architecture is illustrated in Figure~\ref{GritNetArch}. 

\section{Data and Training} 
\label{Data and Training} 
\subsection{Udacity Data}
\label{Udacity Data}
We benchmarked our methods on the student datasets of two Udacity Nanodegree (ND) programs: ND-A and ND-B. These two ND programs were selected specifically because they diverge from each other along many axes. For example, ND-A curriculum has a lower expectation of prior technical knowledge and a relatively higher graduation rate than ND-B. 

In both programs, graduation is defined as completing each of the required projects in a ND program curriculum with a passing grade. When a user officially graduates, their enrollment record is annotated with a time stamp, so it was possible to use the presence of this time stamp as the target label. Users have to graduate before 2017-09-30 to be considered as successfully graduated. 

Each ND program's curriculum contains a mixture of video content, written content, quizzes, and projects. Note that it is not required to interact with every piece of content or complete every quiz to graduate. See below for detailed characteristics of each dataset used for this study.

\begin{itemize}
    \item \textbf{ND-A Dataset}: From the students who enrolled in ND-A program (from 2017-03-07 to 2017-09-30), we selected 1,853 students who had actively engaged with their classroom mentor (believing these to be the students exhibiting full engagement with the curriculum overall). This set of 1,853 students includes 777 students who graduated, yielding a graduation rate of 41.9\%. The length of each student's events streams ranges from 0 to 4,175 events, with an average of 536 events. The curriculum for ND-A program contains 9 projects, 1,025 unique content pages to visit, and 77 quizzes to attempt. 
    
    \item \textbf{ND-B Dataset}: As prescribed above, we selected 8,301 students who actively engaged with their classroom mentor from the students who enrolled in ND-B program (from 2016-06-20 to 2017-09-30). This set of 8,301 students includes 1,005 students who graduated, yielding a graduation rate of 12.1\%. The length of each student's event streams ranges from 1 event to 4,554 events, with an average of 242 events. The curriculum for ND-B program is composed of 6 projects, 668 unique content pages, and 347 quizzes. 
\end{itemize}

For both datasets, an event represents a user taking a specific action (e.g., watching a video, reading a text page, attempting a quiz, or receiving a grade on a project) at a certain time stamp. Some irrelevant data is filtered out during preprocessing, for example, events that occur \textit{before} a user's official enrollment as a result of a free-trial period. It should be noted that no personally identifiable information is included in this data and student equality is determined via opaque unique ids.

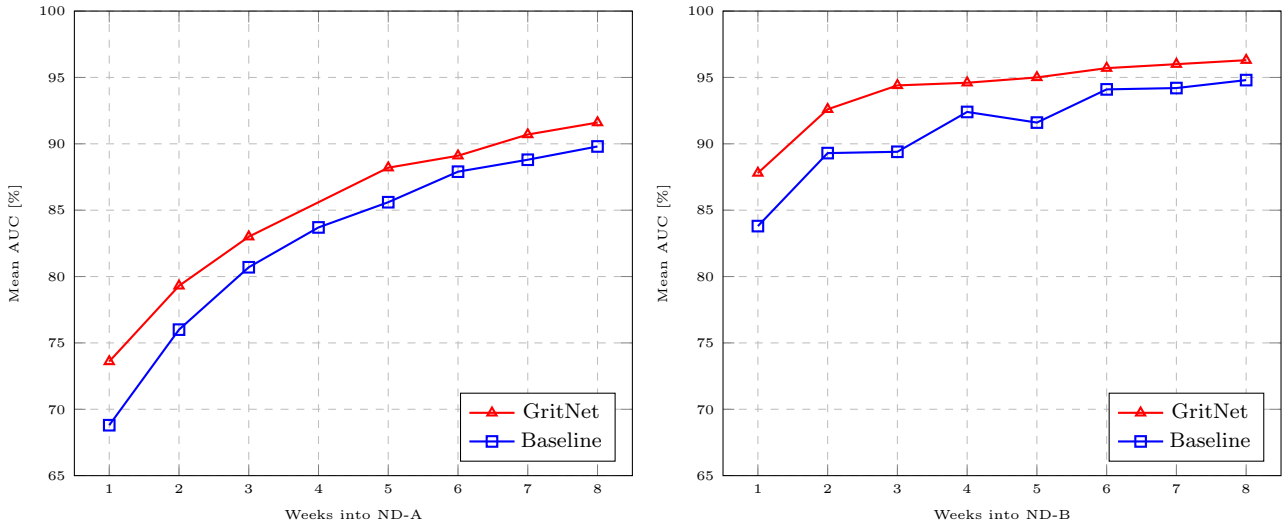
\begin{figure}[ht]
\twocolumn[
\vskip 0.1in
\begin{tikzpicture}

\begin{axis}[
    xlabel={Weeks into ND-A}, 
    ylabel={Mean AUC [\%]},
    xmin=0.5, xmax=8.5, xtick={1,2,...,7,8},
    ymin=65, ymax=100, ytick={65,70,...,95,100},
    xmajorgrids=true, 
    ymajorgrids=true, 
    grid style=dashed,
    label style={font=\tiny}, 
    tick label style={font=\tiny},  
    legend pos=south east,
    legend style={font=\small}
]

\addplot[color=red,mark=triangle,line width=0.8pt]
    coordinates {(1,73.6)
                (2,79.3)
                (3,83.0)
                (5,88.2)
                (6,89.1)
                (7,90.7)
                (8,91.6)};

\addplot[color=blue,mark=square,,line width=0.8pt]
    coordinates {(1,68.8)
                (2,76.0)
                (3,80.7)
                (4,83.7)
                (5,85.6)
                (6,87.9)
                (7,88.8)
                (8,89.8)};

\legend{GritNet, Baseline}
\end{axis}

\end{tikzpicture}
\begin{tikzpicture}

\begin{axis}[
    xlabel={Weeks into ND-B}, 
    ylabel={Mean AUC [\%]},
    xmin=0.5, xmax=8.5, xtick={1,2,...,7,8}, 
    ymin=65, ymax=100, ytick={65,70,...,95,100},
    xmajorgrids=true, 
    ymajorgrids=true, 
    grid style=dashed,
    label style={font=\tiny}, 
    tick label style={font=\tiny},  
    legend pos=south east,
    legend style={font=\small},
]

\addplot[color=red,mark=triangle,line width=0.8pt]
    coordinates {(1,87.8)
                (2,92.6)
                (3,94.4)
                (4,94.6)
                (5,95.0)
                (6,95.7)
                (7,96.0)
                (8,96.3)};

\addplot[color=blue,mark=square,line width=0.8pt]
    coordinates {(1,83.8)
                (2,89.3)
                (3,89.4)
                (4,92.4)
                (5,91.6)
                (6,94.1)
                (7,94.2)
                (8,94.8)};

\legend{GritNet, Baseline}
\end{axis}

\end{tikzpicture}
\caption{Student graduation prediction accuracy comparisons of the GritNet vs baseline models in terms of mean AUC (over all five folds) on two Udacity Nanodegree program datasets: ND-A (left) and ND-B (right). GritNet provides 5.3\% abs (7.7\% rel) accuracy improvements at week 1 for ND-A dataset (left). Notice that for ND-B dataset, the baseline model requires eight weeks of student data to achieve the same performance as the GritNet is able to achieve with only three weeks of student data (right).}
\label{Performance-figure}
\vskip 0.1in
]
\end{figure}

\subsection{Training}
\label{Training}
We learned that the GritNet models are fairly easy to train. The training objective is the negative log likelihood of the observed event sequence of student activities under the model. The binary cross entropy loss is minimized\footnote{In this case, minimizing the binary cross entropy is equivalent to maximizing the log likelihood.} using stochastic gradient descent on mini-batches. 

In our experiment, the BLSTM with forward and backwards LSTM layers containing 128 cell dimensions per direction is used. Embedding layer dimension was grid-searched for the best parameters based on the dataset: from 1024 to 3584 for ND-A set and from 1024 to 5120 for ND-B set. A dropout rate, ranged from 10 to 20\%, applied to the BLSTM output and worked well for both datasets to prevent overfitting during training with a mini-batch size of 32. 

For both baseline and GritNet models, we trained a different model for different weeks, based on students' week-by-week event records, to predict whether each student was likely to graduate. Figure~\ref{Data-figure} shows the number of students and the (longest) event sequence length of a student, both observed at each week.

\section{Prediction Performance} 
\label{Prediction Performance}
\subsection{Evaluation Measure}
\label{Evaluation Measure}
To demonstrate the benefits of the GritNet, we focused on student graduation prediction. Since the true binary target label (1: graduate, 0: not graduate) is imbalanced (i.e., number of 0s outweighs number of 1s), accuracy is not an appropriate metric. Instead, we used the Receiver Operating Characteristic (ROC) for evaluating the quality of the GritNet's predictions. An ROC curve was created by plotting the true positive rate (TPR) against the false positive rate (FPR). In this task, the TPR is the percentage of students who graduate, which the GritNet labels positive, and the FPR is the percent of students who do not graduate, which the GritNet incorrectly labels positive.

The accuracy of each system's prediction was measured by the area under the ROC curve (AUC) which scores between 0 and 100\% (the higher, the better) $-$ with random guess yielding 50\% all the time. We used $5$-fold student level cross-validation, while ensuring each fold contained roughly the same proportions of the two groups (graduate and non graduate) of students.

\subsection{Results}
\label{Results}
For fair comparisons, the baseline performance was optimized by sweeping $\alpha$ values (in Equation~\ref{eq3}) at each week\footnote{The optimized results are quite strong such that initially explored NN models (e.g., DNN, CNN-BLSTM) on the same BoW input features did not yield big win over the baseline.}. The GritNet also required slight hyper-parameter optimization (e.g., embedding dimension as prescribed in Section~\ref{Training}) for the optimal accuracy at each week. 

We have shown that the GritNet really does improve the student graduation prediction accuracy across weeks. From the prediction results on both Udacity datasets in Figure~\ref{Performance-figure}, we clearly see that the performance is similar between the baseline and GritNet models after receiving eight weeks of data about a given student. However, the GritNet is able to achieve significant prediction-quality improvements within the first few weeks of the student experience.

Specifically, the GritNet was able to attain superior performance by more than 5.0\% abs on both ND-A dataset (at week 1) and ND-B dataset (at week 3). Moreover, on ND-B dataset, the baseline model required a wait of two months to reach the prediction accuracy that the GritNet showed within three weeks. We believe this is a crucial advantage of the GritNet, creating a quickly adaptable but accurate metric to estimate long-term student outcomes to accelerate the student feedback loop (which typically takes a few months from enrollment to iterate).

\section{Conclusion}
\label{Conclusion} 
In this paper, we have successfully applied deep learning to the challenging student performance prediction problem which, so far, has not been fully exploited. In contrast to prior work, we formulated the problem as a sequential event prediction problem, introduced a new algorithm called the GritNet to tackle the problem, and demonstrated the superiority of the GritNet using student data from Udacity's Nanodegree programs. 

Two novel properties of the GritNet are that $(1)$ it does not need any feature engineering (it can learn from raw input) and $(2)$ it can operate on any student event data associated with a time stamp (even when highly imbalanced). For future work, we anticipate that incorporating indirect data (e.g., student board activity, interactions with mentors) into the GritNet will potentially further improve the GritNet's impressive performance.

\bibliographystyle{abbrv}
\bibliography{edm}  
 
\balancecolumns
\end{document}